\documentclass[11pt,a4paper]{article}
\usepackage{times,latexsym}
\usepackage{url}
\usepackage[T1]{fontenc}

\usepackage{graphicx}
\usepackage{hyperref}
\usepackage{amsmath}
\usepackage{amsfonts}
\usepackage{enumitem}
\usepackage{multirow}
\usepackage{hhline}
\usepackage{booktabs}
\usepackage{xcolor}
\usepackage{pifont}
\usepackage{dashundergaps}
\usepackage{diagbox}
\newcommand{\cmark}{\ding{51}}%
\newcommand{\xmark}{\ding{55}}%
\usepackage[acceptedWithA]{tacl2018v2}
\usepackage{xspace,mfirstuc,tabulary}

\newif\iftaclinstructions
\taclinstructionsfalse 
\iftaclinstructions
\renewcommand{\confidential}{}
\renewcommand{\anonsubtext}{(No author info supplied here, for consistency with
TACL-submission anonymization requirements)}
\newcommand{\instr}
\fi

\iftaclpubformat 

\else

\fi

\title{Revisiting Negation in Neural Machine Translation}

\author{Gongbo Tang$^1$\quad Philipp R\"onchen$^1$\quad Rico Sennrich$^{2,3}$\quad Joakim Nivre$^1$ \medskip\\
  $^1$Department of Linguistics and Philology, Uppsala University\\
  $^2$Department of Computational Linguistics, University of Zurich\\
  $^3$School of Informatics, University of Edinburgh\\ \medskip
  {\sf firstname.lastname@\{lingfil.uu.se, ed.ac.uk\}}}

\date{}

\begin{document}
\maketitle
\begin{abstract}
In this paper, we evaluate the translation of negation both automatically and manually, in English--German (EN--DE) and English--Chinese (EN--ZH). We show that the ability of neural machine translation (NMT) models to translate negation has improved with deeper and more advanced networks, although the performance varies between language pairs and translation directions. The accuracy of manual evaluation in EN$\rightarrow$DE, DE$\rightarrow$EN, EN$\rightarrow$ZH, and ZH$\rightarrow$EN is 95.7\%, 94.8\%, 93.4\%, and 91.7\%, respectively. 
In addition, we show that under-translation is the most significant error type in NMT, which contrasts with the more diverse error profile previously observed for statistical machine translation.
To better understand the root of the under-translation of negation, we study the model's information flow and training data.
While our information flow analysis does not reveal any deficiencies that could be used to detect or fix the under-translation of negation, we find that negation is often rephrased during training, which could make it more difficult for the model to learn a reliable link between source and target negation.
We finally conduct intrinsic analysis and extrinsic probing tasks on negation, showing that NMT models can distinguish negation and non-negation tokens very well and encode a lot of information about negation in hidden states but nevertheless leave room for improvement.
\end{abstract}

\section{Introduction}
\label{sec:intro}

Negation is an important linguistic phenomenon in machine translation, as errors in translating negation may change the meaning of source sentences completely. There are many studies on negation in statistical machine translation (SMT) \cite{collins2005clause,li2009chinese,wetzel2012enriching,baker2012modality,fancellu2014applying,fancellu2015negation}, but studies on negation in neural machine translation (NMT) are quite limited and results are partly conflicting. 
For example, \newcite{bentivogli2016neural} find that negation is still challenging, whereas \newcite{bojar2018final} show that NMT models almost make no mistakes on negation using 130 sentences with negation from three language pairs as the evaluation set. Hence, it is still not clear how well NMT models perform on the translation of negation. 

In this paper, we present both automatic and manual evaluation of negation in NMT, in English--German (EN--DE) and English--Chinese (EN--ZH). The automatic evaluation is based on contrastive translation pairs and studies translation from English into German/Chinese (EN$\rightarrow$DE/ZH). The manual evaluation targets translation in all four translation directions. 
We find that the modeling of negation in NMT has improved with deeper and more advanced networks. The contrastive evaluation shows that deleting negation from references is more confusing to NMT models compared to inserting negation into references. 
For the manual evaluation, NMT models make fewer mistakes on negation in EN--DE, than in EN--ZH, and there are more errors on negation in DE/ZH$\rightarrow$EN than in EN$\rightarrow$DE/ZH. Moreover, under-translation is the most prominent error type in three out of four directions. 

The black-box nature of neural networks makes it hard to interpret how NMT models handle the translation of negation. In \newcite{ding2017visualizing}, neither attention weights nor layer-wise relevance propagation (LRP) can explain why negation is under-translated. We are interested in whether the information about negation is not well passed to the decoder. Thus, we investigate the negation information flow in NMT models by raw attention weights and attention flow \cite{abnar2020quantifying}. 
We demonstrate that the under-translation of cues is not caused simply by a lack of negation information transferred to the decoder. 
We further explore the mismatch between source and target sentences --- negation cues appearing only on the source side or only on the target side. We find that there are roughly 17.4\% mismatches in the training data in ZH--EN. These mismatches could confuse NMT models and make the learning harder. We suggest to distill or filter training data by removing the sentence pairs with mismatches to make the learning easier. 
In addition, we conduct intrinsic analysis and extrinsic probing tasks, to explore how much information about negation has been learned by NMT models. The intrinsic analysis based on cosine similarity shows that NMT models can distinguish negation and non-negation tokens very well. The probing results on negation detection reveal that NMT can encode a lot of information about negation in hidden states but still leaves much room for improvement. Moreover, encoder hidden states capture more information about negation than decoder hidden states.

\section{Related Work}

\subsection{Negation in MT}

\newcite{fancellu2015negation} conduct a detailed manual error analysis and consider three categories of errors, \textit{deletion}, \textit{insertion}, and \textit{reordering}. They find that negation scope is most challenging and \textit{reordering} is the most frequent error type in SMT. Here we study the performance of NMT models on translating negation. 

\newcite{bentivogli2016neural} and \newcite{beyer2017can} find that NMT is superior to SMT in translating negation. \newcite{bentivogli2016neural} observe that placing the German negation cue \textit{nicht} correctly during translation is a challenge for NMT models, which is determined by the focus of negation and need to detect the focus correctly. \newcite{bojar2018final} evaluate MT models on negation, translating from English into Czech, German, and Polish, using 61, 36, 33 sentences with negation as the test sets. They find that NMT models almost make no mistakes on negation compared to SMT -- NMT models only make two mistakes in the English--Czech test set. In this paper, we will conduct manual evaluation on four directions with larger evaluation sets, to get a more comprehensive picture of the performance on translating negation. 

\newcite{sennrich2017grammatical} evaluates subword-level and character-level NMT models on the \textit{polarity} set of \textit{LingEval97} and finds that negation is still a challenge for NMT, via scoring contrastive translation pairs. More specifically, the deletion of negation cues causes more errors. \newcite{ataman2019importance} show that character-level models perform better than subword-level models on negation. Instead, we evaluate NMT models with different neural networks to learn their abilities to translate negation, by scoring contrastive translate pairs. 

\newcite{ding2017visualizing} find that neither attention weights nor LRP can explain under-translation errors on a negation instance. Thus understanding the mechanism of dealing with negation is still a challenge for NMT. 
Most recently, \newcite{hossain2020non} study the translation of negation on 17 translation directions. They show that negation is still a challenge to NMT models and find that there are fewer negation related errors when the language is similar to English, with respect to the typology of negation. 
In our work, we conduct both automatic and manual evaluation on negation, and explore the information flow of negation to answer whether under-translation errors are caused by a lack of negation information transferred to the decoder. 

\subsection{Negation in Other Areas of NLP}

Negation projection is the task of projecting negations from one language to another language, which can alleviate the workload of annotating negation. \newcite{liu2018negpar} find that using word alignment to project negation does not help the annotation process. They also provide the \textit{NegPar} corpus, an EN--ZH parallel corpus annotated for negation. Here we apply probing classifiers to directly generate negation annotations on Chinese using hidden states. 

Negation detection is the task of recognizing negation tokens, which can estimate the ability of a model to learn negation. 
\newcite{fancellu2018neural} utilize LSTMs, dependency LSTMs, and graph convolutional networks (GCN) to detect negation scope, using part-of-speech tags, dependency tags, negation cues as features. Recently the pre-trained contextualized representations have been widely used in various NLP tasks. \newcite{khandelwal2019negbert} employ BERT \cite{devlin2019bert} for negation detection, including negation cue detection, scope detection and event detection. \newcite{sergeeva2019neural} apply ELMo \cite{peters2018deep} and BERT to negation scope detection and achieve new state-of-the-art results on two negation data sets. Instead of pursuing better results, here we aim to probe how much information about negation has been encoded in hidden states in a negation detection task.

\section{Background}

\subsection{Negation}

Negation in text generally has four components: cues, events, scope, and focuses. The cues are the words expressing negation. An event is the lexical component that a cue directly refers to. The scope is the part of the meaning that is negated and the focus is the most explicitly negated part of the scope \cite{huddleston2002cambridge,morante2012conandoyle}.

\textit{NegPar} is a parallel EN--ZH corpus annotated for negation. The English part is based on \textit{ConanDoyle-neg} \cite{morante2012conandoyle}, a collection of four Sherlock Holmes stories. Some scope-related phenomena are re-annotated for consistency. The annotations are extended onto its Chinese translations. Here are two annotation examples: 

\indent \indent English:  
\underline{There was} \textbf{no} \framebox{\underline{response}}. \\  
\indent \indent Chinese: 
\textbf{mei} \underline{you ren} \framebox{\underline{da ying}}. \\
\indent \indent \indent \indent \indent \ \ \ 
(no have people answer reply.)

\noindent
In these examples, \textit{\textbf{no}} and \textit{\textbf{mei}} marked in bold are the cues; \textit{response} and \textit{da ying} enclosed in boxes are the events; the underlined words belong to the negation scope. In \textit{NegPar}, negation events are subsets of negation scope, and negation focuses are not annotated. 
Table~\ref{table-negpar} shows detailed statistics of \textit{NegPar}. Note that a negation instance may not have all the three components. Moreover, not all parallel sentence pairs have negation in both source and target sentences. 
For more details, please refer to \newcite{liu2018negpar}. 

Due to the lack of parallel data annotated for negation, most of the negated sentences in the previous studies are selected randomly. In \textit{NegPar}, not only negation cues, but also events and scope are annotated which is beneficial to evaluating NMT models on negation and exploring the ability of NMT models to translate negation.

\subsection{Contrastive Translation Pairs} 
  \label{sub:contrastive_evaluation}

\begin{table}[htbp]
\begin{center}
\scalebox{0.96}{
\begin{tabular}{cccccc}
\toprule
\ &\ &Train&Dev&Test&Total \\
\cmidrule(l){1-6}
\multirow{3}{*}{English} &Cue & 984 & 173 & 264 & 1,421 \\
 &Event & 616 & 122 & 173 & \phantom{0,}911 \\
 &Scope & 887 & 168 & 249 & 1,304 \\
\addlinespace
\multirow{3}{*}{Chinese}&Cue & 1,209 & 231 & 339 & 1,779 \\
 &Event& \phantom{0,}756 & 163 & 250 & 1,169 \\
 &Scope& 1,160 & 227 & 338 & 1,725 \\
\bottomrule
\end{tabular}}
\caption{\label{table-negpar} Statistics of negation components in \textit{NegPar}.}
\end{center}
\end{table}

\begin{table}[htbp]
\begin{center}
\scalebox{0.83}{
\begin{tabular}{ll}
\toprule
Deletion &Insertion \\
\midrule
deleting \textit{nicht} (not) &inserting \textit{nicht}\\
replacing \textit{kein} (no) with \textit{ein} (a) &replacing \textit{ein} with \textit{kein}\\
deleting \textit{un-} &inserting \textit{un-}\\
\addlinespace
\bottomrule
\end{tabular}}
\caption{\label{table-polarity} Six ways to reverse the polarity of sentences from the \textit{polarity} category of \textit{LingEval97}.}
\end{center}
\end{table}

Since we evaluate NMT models explicitly on negation, BLEU \cite{papineni2002bleu} as a metric of measuring overall translation quality is not helpful. We conduct the targeted evaluation with contrastive test sets in which human reference translations are paired with one or more contrastive variants, where a specific type of error is introduced automatically. 

NMT models are conditional language models that assign a probability $P(T|S)$ to a given source sentence $S$ and the target sentence $T$. If a model assigns a higher probability to the correct target sentence than to a contrastive variant that contains an error, we consider it as a correct decision. The accuracy of a model on such a test set is the percentage of cases where the correct target sentence is scored higher than all contrastive variants. 

\textit{LingEval97} \cite{sennrich2017grammatical} has over 97,000 EN$\rightarrow$DE contrastive translation pairs featuring different linguistic phenomena. 
In this paper, we focus on the \textit{polarity} category which is related to negation and consists of 26,803 instances. For contrastive variants, the polarity of translations are reversed by inserting or deleting negation cues. Table~\ref{table-polarity} illustrates how the polarity is reversed.

\subsection{Attention Flow}

In Transformer models, the hidden state of each token is getting more contextualized as we move to higher layers. Thus, the raw attention weights are not the actual attention to the input tokens.

Recently, \newcite{abnar2020quantifying} have proposed attention flow to approximate the information flow. Attention flow considers not only the attention weights to the previous layer but also to all the lower layers. 
Formally, in the self-attention networks, given a directed graph $G = (V, E)$, where $V$ is the set of nodes, and $E$ is the set of edges; each hidden state or word embedding from different layers is a node; the attention weight is the value of an edge. Given a source node $s$ and a target node $t$, the attention flow is the flow of edges between $s$ and $t$, where the flow value should not exceed the capacity of each edge and input flow should be equal to output flow for the intermediate nodes in the path $s$ to $t$. They apply a maximum flow algorithm to find the flow between $s$ and $t$ in a flow network. 

In short, the attention flow utilizes the minimum value of the attention weights in each path, and also employs the residual connections of attention weights. 
They find that the patterns of attention flow get more distinctive in higher layers compared to the raw attention. 
Moreover, attention flow yields higher correlations with the importance scores of input tokens obtained by the input gradients, compared to using the raw attention weights. 
\newcite{abnar2020quantifying} explore the attention flow of the encoder self-attention in the case of pre-trained language models. Here we compute the attention flow from decoder layers to source word embeddings, in the context of NMT.

\section{Evaluation}
In this section, we present the results of both automatic and manual evaluation on negation in EN--DE and EN--ZH, to get a more comprehensive picture of the performance on translating negation. 

\subsection{NMT Models}

We use the \textit{Sockeye} \cite{Hieber2017sockeye} toolkit to train NMT models.  For EN$\rightarrow$DE, we train RNN-, CNN-, and Transformer-based models, following the settings provided by \newcite{tang2018why}. For the other directions, we only train Transformer models. Table~\ref{table-settings} shows the more detailed settings.

\begin{table}[htbp]
\scalebox{0.75}{
\begin{tabular}{ll}
\toprule
Neural network depth & 8/6 (EN--DE/ZH) \\ 
Kernel size of CNNs & 3 \\ 
Trans. Att. head & 8 \\ 
Learning rate (initial) & 2e-04 \\ 
Embedding\&hidden unit size & 512 \\ 
Mini-batch size (token) & 4,096 \\ 
Dropout (Trans./RNN\&CNN) &  0.1/0.2 \\ 
RNN encoder & 1 biLSTM + 6 uniLSTM  \\ 
Optimizer & \textit{Adam} \citep{Kingma2014AdamAM} \\ 
Checkpoint frequency  &  4,000  \\ 
Label smoothing  & 0.1 \\ 
Early stopping & 32 \\
\bottomrule
\end{tabular}}
\caption{\label{table-settings} Settings for training NMT models. }
\end{table}

\begin{table}[htbp]
\begin{center}
\scalebox{0.81}{
\begin{tabular}{cccccc}
\toprule
\multicolumn{3}{c}{EN$\rightarrow$DE}& DE$\rightarrow$EN &EN$\rightarrow$ZH &ZH$\rightarrow$EN \\
\cmidrule(l){1-3} \cmidrule(l){5-5} \cmidrule(l){6-6} \cmidrule(l){4-4}
RNN&CNN&Trans.&Trans.&Trans.&Trans. \\
\cmidrule(l){1-6}
25.2& 25.3 & 27.6& 34.3 & 33.9 & 23.5\\
\bottomrule
\end{tabular}}
\caption{\label{table-bleu} BLEU scores of NMT models with different architectures on the test sets (\textit{newstest2017}). \textit{Trans.} is short for \textit{Transformer}. }
\end{center}
\end{table}

The training data is from the WMT17 shared task \cite{wmt17}.\footnote{\url{http://www.statmt.org/wmt17/translation-task.html}} There are about 5.9 million and 24.7 million sentence pairs in the training set of EN--DE and EN--ZH, respectively, after preprocessing with Moses scripts. Note that the training data on EN--ZH is from the official preprocessed data.\footnote{\url{http://data.statmt.org/wmt18/translation-task/preprocessed/zh-en/}} 
The Chinese segmentation is based on Jieba.\footnote{\url{https://github.com/fxsjy/jieba}} 
We learn a joint BPE model with 32K subword units \cite{sennrich16sub} for EN--DE, and two BPE models with 32K subword units for Chinese and English, respectively. We employ the single model that has the best perplexity on the validation set for the evaluation, without any ensembles. 
Table~\ref{table-bleu} shows the BLEU scores of the trained NMT models on \textit{newstest2017}, which are computed by \textit{sacrebleu} \cite{post2018sacre}.\footnote{\url{https://github.com/mjpost/sacrebleu}}

Since these NMT models are trained with single sentences, feeding an input with multiple sentences into these models is likely to get an incomplete translation. To avoid these errors, we feed the sentence with negation cues into NMT models individually for the manual evaluation.

\begin{figure*}[htbp]
\centering
        \includegraphics[totalheight=4cm]{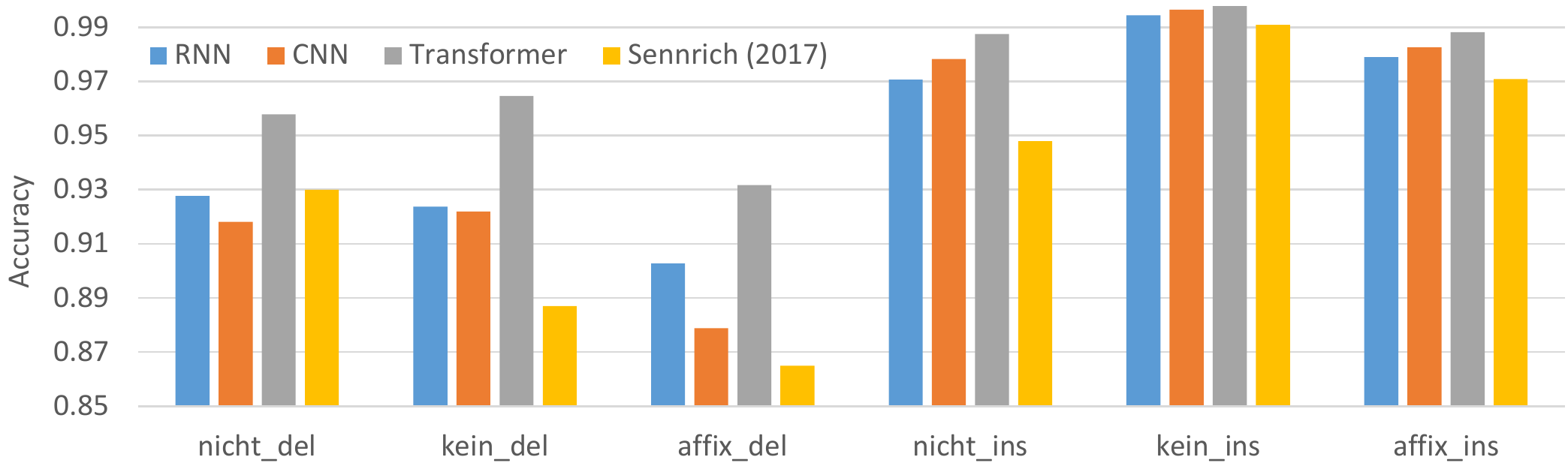}
    \caption{Performance of NMT models on scoring contrastive translations, in EN$\rightarrow$DE, using the \textit{polarity} category of \textit{LingEval97}. The first three groups are on negation deletion, deleting \textit{nicht}, \textit{kein} and affixes, while the last three groups are on negation insertion. }
    \label{fig:eval-lingeval97}
\end{figure*}

\subsection{Automatic Evaluation}

For the automatic evaluation, we let NMT models score contrastive translation pairs, in EN$\rightarrow$DE and EN$\rightarrow$ZH. 

\subsubsection{EN$\rightarrow$DE}

\newcite{sennrich2017grammatical} has evaluated subword-level and character-level RNN-based models. Here we evaluate NMT models with different architectures, RNN-, CNN-, and Transformer-based models. The test set is the \textit{polarity} category of \textit{LingEval97}. Figure~\ref{fig:eval-lingeval97} displays the accuracy of NMT models. 

Our NMT models are superior to the models in \newcite{sennrich2017grammatical}, except that \textit{CNN} is inferior in the group \textit{nicht\_del}. Generally, we see that the performance on negation is getting better with the evolution of NMT models, with the \textit{Transformer} consistently scoring best, and substantially better (by up to 8 percentage points) than the shallow RNN \cite{sennrich2017grammatical}. The accuracy of the \textit{Transformer} varies from 93.2\% to 99.8\%, depending on the group, which we consider quite strong.

It is interesting that NMT models make fewer mistakes when inserting negation cues into the reference compared to deleting negation cues from the reference, which means that positive contrastive variants are more confusing to NMT models. 
This is consistent with the results in \newcite{fancellu2015negation}, that SMT models make more errors when generating positive sentences than generating negative sentences, in terms of insertion/deletion errors. We will explore under-translation errors in the following sections. 

\subsubsection{EN$\rightarrow$ZH}

Following the \textit{polarity} category in \textit{LingEval97}, we create a contrastive evaluation set for negation on EN$\rightarrow$ZH, using the development and test sets from the WMT shared translation task 2017--2020.\footnote{\url{https://github.com/tanggongbo/negation-evaluation-nmt}} The contrastive evaluation set also has two sub-categories: negation deletion and negation insertion. 
We first select the five most popular Chinese negation cues -- ``bu'', ``mei'', ``wu'', ``fei'', and ``bie''. Then, we manually delete the negation cue from the reference or insert a negation cue into the reference, without affecting the grammaticality. The negation deletion and negation insertion categories have 2,005 and 3,062 instances with contrastive translations, respectively. 

As Transformer models are superior to RNN- and CNN-based models, here we only evaluate Transformer models. The accuracy on negation deletion and negation insertion categories is 92.1\% and 99.0\%, respectively. 
We can see that Transformer models perform quite well on EN$\rightarrow$ZH, but not as well as on EN$\rightarrow$DE. 
In accord with the finding in EN$\rightarrow$DE, Transformer models here in EN$\rightarrow$ZH also perform worse on the negation deletion category. 

\begin{table*}[htbp]
\begin{center}
\scalebox{0.95}{
\begin{tabular}{ll}
\toprule
Category & Description \\
\midrule
\textit{Correct}&cues are translated into cues correctly\\
\textit{Rephrased}&cues are translated correctly but not into a cue\\
\textit{Reordered}& cues are translated but modify wrong constituents (incorrect scope/focus) \\
\textit{Incorrect}&cues are translated but the event is translated incorrectly or the meaning is reversed \\
\textit{Dropped}&cues are not translated at all\\
\addlinespace
\bottomrule
\end{tabular}}
\caption{\label{table-group-trans} Descriptions of the five translation categories.}
\end{center}
\end{table*}

\begin{table*}[ht!]
\begin{center}
\scalebox{0.95}{
\begin{tabular}{lcccccc}
\toprule
\ &\textit{Correct}&\textit{Rephrased}&\textit{Reordered}&\textit{Incorrect}&\textit{Dropped} &\textit{Accuracy}\\
\midrule  
EN$\rightarrow$DE&258 (92.8\%)&\phantom{0}8 (\phantom{0}2.9\%)&2 (0.7\%)&\phantom{0}3 (1.1\%)&\phantom{0}7 (2.5\%) & 95.7\%\\ 
DE$\rightarrow$EN &232 (92.8\%)&\phantom{0}5 (\phantom{0}2.0\%)&2 (0.8\%)&11 (4.4\%)&\phantom{0}0 (0.0\%) & 94.8\% \\ 
\addlinespace
EN$\rightarrow$ZH&393 (90.0\%)&15 (\phantom{0}3.4\%)&3 (0.7\%)&10 (2.3\%)&16 (3.7\%)& 93.4\%\\ 
ZH$\rightarrow$EN &451 (80.1\%)&65 (11.6\%)&3 (0.5\%)&21 (3.7\%)&23 (4.1\%)& 91.7\%\\ 
\bottomrule
\end{tabular}}
\caption{\label{table-eval-manual} Manual evaluation results in EN--DE and EN--ZH. Accuracy is the sum of \textit{correct} and \textit{rephrased}. }
\end{center}
\end{table*}

\subsection{Manual Evaluation}

We have evaluated NMT models on negation with contrastive translation pairs. However, scoring contrastive translation pairs is not the same as evaluating the translations directly. The contrastive translations only insert or delete a negation cue compared to the references, which is quite different from the generation of NMT models. In addition, the automatic evaluation only gives us the general performance on negation without any details on how negation is translated. 
Thus, we further conduct manual evaluation on EN--DE and EN--ZH. 

Due to the lack of parallel data annotated for negation, most of the negated sentences in previous studies have no annotations and are selected randomly. 
In \textit{NegPar}, not only negation cues, but also events and scope are annotated, which is beneficial for evaluating NMT models on negation and exploring the ability of NMT models to learn negation. 
These annotations allow us to evaluate negation from the perspectives of cues, events, and scope, rather than negation cues only. 
Thus, for EN--ZH, we conduct the manual evaluation based on \textit{NegPar}, using both the development set and the test set. 
For EN--DE, we evaluate 250 sentences with negation cues that are randomly selected from \textit{LingEval97} in each direction. 

Given the strong performance of Transformer models in the automatic evaluation, we focus on this architecture for the manual evaluation. 
We classify the translations of negation into five categories: \textit{Correct}, \textit{Rephrased}, \textit{Reordered}, \textit{Incorrect}, and \textit{Dropped}, depending on whether the cue, event and the scope are translated correctly. More detailed descriptions are provided in Table~\ref{table-group-trans}.

Table~\ref{table-eval-manual} gives the absolute frequency and percentage of each translation category in all the translation directions.\footnote{\url{https://github.com/tanggongbo/negation-evaluation-nmt} provides the details.}  The accuracy of translating negation is the sum of \textit{correct} and \textit{rephrased}, and the accuracy in EN$\rightarrow$DE, DE$\rightarrow$EN, EN$\rightarrow$ZH, and ZH$\rightarrow$EN is 95.7\%, 94.8\%, 93.4\%, and 91.7\%, respectively. We can see that NMT models perform better at translating negation in DE--EN than in ZH--EN. 
In addition, under-translation errors are the main errors in three out of four directions while reordering errors only account for less than 1\% in all directions. This contrasts with the results reported for SMT by \citet{fancellu2015negation}, where reordering was a more severe problem than under-translation. It is reasonable because NMT models are conditional language models, and have fewer word order errors, compared to SMT models \cite{bentivogli2016neural}, thus there are fewer reordering errors on translating negation. We can tell that the main error types with respect to negation have shifted from SMT to NMT.

\subsubsection{EN--DE}
As Table~\ref{table-eval-manual} shows, most of the translations belong to \textit{correct}. The accuracy in EN$\rightarrow$DE is 0.9\% greater than that in DE$\rightarrow$EN. 2.5\% negation cues are not translated in EN$\rightarrow$DE, while all the negation cues are translated by NMT models in DE$\rightarrow$EN. However, there are more sentences where the negation events are not translated correctly in DE$\rightarrow$EN. 
Compared to \newcite{bojar2018final}, our evaluation results for EN-DE are 4.3\% lower. One possible reason for the difference is that our evaluation is based on a larger data set; another possible reason is that we also consider the translation of negation events and scope.

\subsubsection{EN--ZH}

Similar to the results in EN--DE, the accuracy in translating from English is greater than in translating into English. The accuracy in ZH$\rightarrow$EN is 1.7\% lower than in EN$\rightarrow$ZH. There are more instances of negation that are rephrased in the translations in ZH$\rightarrow$EN, without any negation cues in the translations. The NMT model in ZH$\rightarrow$EN also makes more under-translation errors.

\begin{table*}[htbp]
\begin{center}
\scalebox{0.8}{
\begin{tabular}{llll}
\toprule
Category&Source&Translation&Reference\\
\midrule
\shortstack[l]{\textit{Correct} \\ \ }&\shortstack[l]{\underline{would do him} \textbf{no} \underline{harm} \\ \ }&\shortstack[l]{\dashuline{bu} hui shang hai ta\\ (not able to harm him)}&\shortstack[l]{ dui ta bu hui you shen me hai chu\\ (to him no able have any harm) }\\
\midrule
\shortstack[l]{\textit{Rephrased} \\ \ }&\shortstack[l]{\textbf{bu} \underline{xi fei yong}  \\ (no spare expense use)}&able to \dashuline{spend enough} money & spare no expense \\
\midrule
\shortstack[l]{\textit{Reordered} \\ \ }&\shortstack[l]{\underline{yi ge xing qi} \textbf{bu} \underline{jian mian}\\ (a week no meet) }&\uwave{no one} could meet for a week &be invisible for a week \\
\midrule
\shortstack[l]{\textit{Incorrect} \\ \ }&\shortstack[l]{\underline{spare} \textbf{no} \underline{expense} \\ \ }&\shortstack[l]{\uwave{bu yao hua qian} mai\\ (not spend money to buy)} &\shortstack[l]{bu xi fei yong \\ (not spare expense)} \\
\midrule
\shortstack[l]{\textit{Dropped} \\ \ }&\shortstack[l]{\textbf{bu} \framebox{\underline{xing}}, \underline{Mo li luo zhi dao le} \\ (not fortunate, Murillo know truth already)}&\uwave{fortunately}, Murillo knew that &Unhappily , Murillo heard of \\
\bottomrule
\end{tabular}}
\caption{\label{table-examples} Translation examples (segments) from different categories. These segments are a subset of negation scope. The word in bold in the source is the cue. Words with dashed lines below are correct translations and words with wavy lines below are incorrect translations. }
\end{center}
\end{table*}

Table~\ref{table-examples} further provides some translation examples. 
In the category \textit{Rephrased}, negation cues are not directly translated into negation cues. Instead, the negation is paraphrased in a positive translation. In the \textit{Rephrased} example, although there is no cue in the translation, the meaning is paraphrased by translating \textit{bu xi (no spare)} into \textit{spend}. 
In the \textit{Reordered} example, the cue \textit{bu} in the source is supposed to modify \textit{jian} (meet), but the translation of the cue is placed before \textit{one}, modifying the subject \textit{one} instead of \textit{meet}. 
In addition, even though the negation cues are translated, the negation events could be translated incorrectly, which can also have a severe impact on the translation. For the fourth example, there is a cue in the translation but \textit{spare} in the source is translated into \textit{spend}, which reverses the meaning completely. 
For the last example, the cue \textit{bu (no)} is skipped and only the event \textit{xing (fortunate)} gets translated. 

We further check the under-translation errors of negation cues and find that some of them are caused by multi-word expressions (idioms), especially when translating Chinese into English. For example, \textit{wu} (no) in \textit{wu\_bing\_shen\_yin} (no disease groan cry) is not translated. \newcite{fancellu2015negation} have shown that the cues will not be under-translated if they are separate units in SMT. Thus, these words are then segmented into separate characters and the input is fed into NMT models again. This does fix a few errors. The \textit{wu} (no) in \textit{wu\_bing\_shen\_yin} gets translated but the second \textit{bu} (not) in \textit{bu\_gao\_bu\_ai} (not tall not short) is still not translated. Note that we only changed the segmentation during inference which is sub-optimal. We aim to show that the segmentation also could cause under-translation errors.

\section{Interpretation} 
\label{sec:Interpretation}

There are few studies on interpreting NMT models with respect to negation. Since Table~\ref{table-eval-manual} has shown that NMT models in EN--ZH suffer from more errors on negation, and since \textit{NegPar} provides annotations of negation, we focus on interpreting NMT models in EN--ZH. 
NMT models consist of several components and we are interested in the information flow of negation to answer whether the under-translation is caused by not passing enough negation information to decoders, as well as exploring the ability of NMT models to learn negation. 

\subsection{Under-Translation Errors}

Under-translation is the most frequent error type in our evaluation. If a negation cue is not translated by NMT models, either the negation information is not passed to the decoder properly, or the decoder does not utilize such information for negation generation. We employ raw attention weights and attention flow to explore the information flow.

\subsubsection{Attention Distribution}

Encoder-decoder attention weights can be viewed as the degree of contribution to the current word prediction. They have been utilized to locate unknown words and to estimate the confidence of translations \citep{jean2015using,gulcehre2016pointing,Rikters2017confidence}. However, previous studies have found that attention weights cannot explain the under-translation of negation cues \citep{ding2017visualizing}. 
In this section, we first focus on the under-translated negation cues, checking the negation information that is passed to the decoder by the encoder-decoder attention. We compare the attention weights paid to negation cues, when they are under-translated and when they are translated into reference translations. 

We extract attention distributions from each attention layer when translating sentences from the development set. Each attention layer has multiple heads and we average\footnote{We also used maximum weights to avoid misleading conclusions when using average weights if the negation is modeled by a specific head, and we got the same conclusion.} the attention weights from all the heads. We utilize constrained decoding \citep{Post2018fast} to generate reference translations to get gold attention distribution. 
We find that source negation cues attract much less attention compared to when they are translated into references. 
Thus, we hypothesize that sufficient information about negation has not been passed to the decoder, and we can utilize the attention distribution to detect under-translated cues. 

Now we further explore the attention distribution of under-translated and correctly translated cues, without using the gold attention distribution. We compute the Spearman correlation ($\rho$) between the weights and categories. If $|\rho|$ is close to $1$, then categories have a high correlation with attention weights. 
However, the largest $|\rho|$ in EN$\rightarrow$ZH and ZH$\rightarrow$EN is 0.15 and 0.23, respectively, which means that there is almost no correlation between attention weights and categories. 
We inspect the weights and find that the weights to correctly translated cues range from 0.01 to 0.68, which cover most of the weights to dropped cues. This means that we cannot detect under-translated cues by raw attention weights. 

As raw attention weights in Transformer are not the actual attention to input tokens, in the next section, we will apply attention flow, which has been shown to have higher correlation with the input gradients, to measure the negation flow.

\subsubsection{Attention Flow}

We compute the attention flow to negation cues belonging to different groups; the input nodes are the hidden states from decoder layers; the output node is the word embedding of the negation cue. 
We utilize the maximum attention flow from the decoder to represent the attention flow to each source cue, and report the average value of all the attention flow. 
Table~\ref{table-attflow} shows the attention flow values from different decoder layers to source cues, and the absolute value of Spearman correlation ($\rho$) between attention flow and the cue's category. 
The attention flow values range from 0.70 to 0.91 for all the cues, which means that most of the cue information has been passed to the decoder and that the under-translation is not caused by not passing negation information to the decoder. 

\begin{table}[htbp]
\begin{center}
\scalebox{0.72}{
\begin{tabular}{cccccc}
\toprule
& \ & \multicolumn{2}{c}{EN$\rightarrow$ZH }& \multicolumn{2}{c}{ZH$\rightarrow$EN} \\
\cmidrule(l){3-4}  \cmidrule(l){5-6}
Layer & Group &Attention flow&|$\rho|$&Attention flow&$|\rho|$ \\
\midrule
\multirow{2}{*}{2}&\cmark&0.89&\multirow{2}{*}{0.04}&0.80&\multirow{2}{*}{0.15}\\
&\xmark&0.90&&0.70\\
\midrule
\multirow{2}{*}{4}&\cmark&0.89&\multirow{2}{*}{0.06}&0.85&\multirow{2}{*}{0.08}\\
&\xmark&0.91&&0.84\\
\midrule
\multirow{2}{*}{6}&\cmark&0.77&\multirow{2}{*}{0.06}&0.82&\multirow{2}{*}{0.07}\\
&\xmark&0.78&&0.72\\
\bottomrule
\end{tabular}}
\caption{\label{table-attflow} 
Attention flow values from different decoder layers to source cues, and the absolute value of Spearman correlation ($\rho$) between attention flow and the cue's category. 
\cmark \ represents the correctly translated cues and \xmark \ represents the under-translated cues. }
\end{center}
\end{table}

In addition, the attention flow values in \textit{Dropped} and \textit{Correct} are almost the same in EN$\rightarrow$ZH and the correlation is smaller than 0.1. In ZH$\rightarrow$EN, the attention flow is more distinct in the two cue groups, but the correlation values are still smaller than 0.15. 
Compared to raw attention weights, attention flow can provide more accurate information flow to the decoder, but neither raw attention weights nor attention flow exhibit any correlation between under-translation and the amount of negation information passed to the decoder. 

Our analysis indicates that under-translation of negation cues may still occur even though there is information flow from the source negation cue to the decoder. 
This indicates that methods to manipulate the attention flow, such as coverage models or context gates \citep{tu2016modeling,tu2017context} may not be sufficient to force the model to produce negation cues.
Our results also indicate that under-translation of negation cues may not be easily detectable via an analysis of attention.

\subsubsection{Training Data Considerations}

\begin{figure*}[htbp]
\centering
        \includegraphics[totalheight=4.6cm]{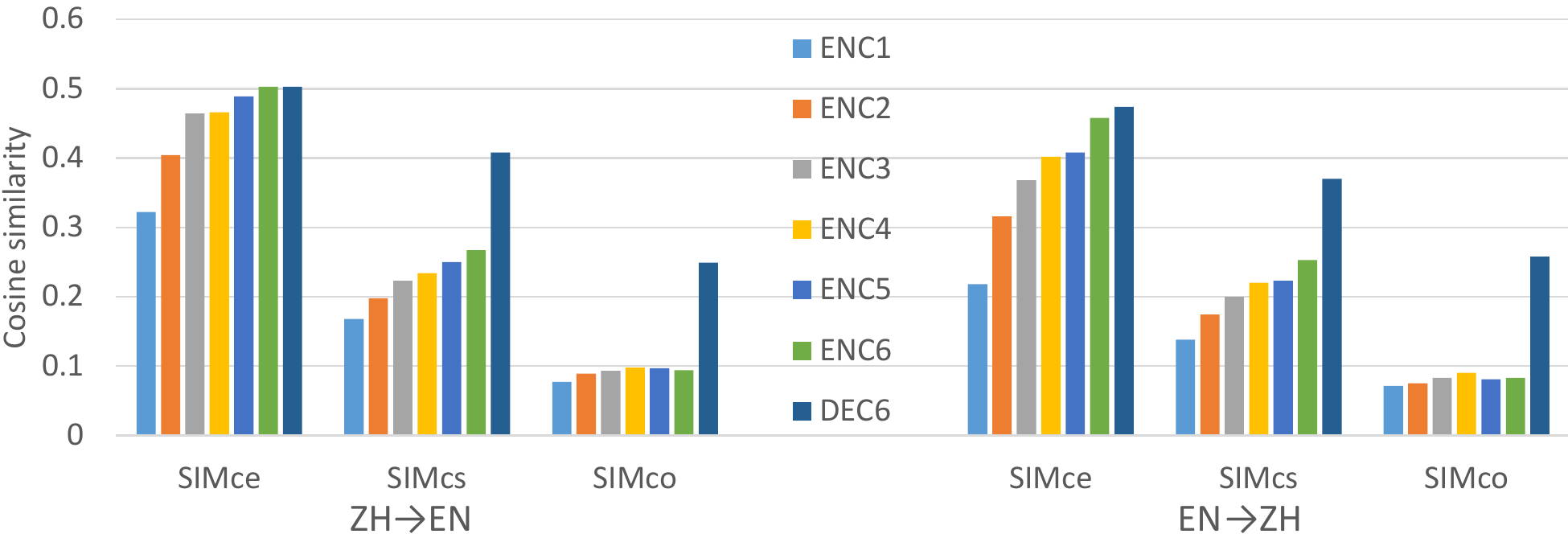}
    \caption{Cosine similarity between negation cues and events, scope, and non-negation tokens in ZH--EN, using hidden states from different layers. \textit{ENC$i$} represents hidden states from the $i$th encoder layer and \textit{DEC6} denotes hidden states from the 6th decoder layer. }
    \label{fig:similarity}
\end{figure*}  

To further investigate why a model would fail to learn the seemingly simple correspondence (in the language pairs under consideration) between source and target side negation cues, we turn to an analysis of the parallel training data.
Our manual analysis of the test sets has shown a sizeable amount (2--11\%) of rephrasing where the translation of a negation is correct, but avoids grammatical negation.
We hypothesize that such training examples could weaken the link between grammatical negation cues in the source and target, and favour their under-translation.

\begin{table}[htbp]
\begin{center}
\scalebox{0.92}{
\begin{tabular}{ccc}
\toprule
\diagbox{EN}{ZH}& has\_cue & no\_cue\\
\midrule
has\_cue &2.60M (10.5\%) & \textbf{\phantom{2}0.15M (\phantom{7}0.6\%)} \\
no\_cue & \textbf{4.16M (16.8\%)}& 17.84M (72.1\%) \\
\bottomrule
\end{tabular}}
\caption{\label{table-cue-mismatch} 
Statistics of sentence pairs with and without cues in ZH--EN, including absolute number and ratio. ``M'' is short for million. Numbers in bold denote sentence pairs with cue-mismatch. }
\end{center}
\end{table}

We perform an automatic estimate of cue-matches and cue-mismatches between source and target in the training data based on a short list of negation words.\footnote{English negation words: \textit{no, non, not, 't, nothing, without, none, never, neither}. Chinese negation characters: \textit{bu, mei, wu, fei, bie, wei, fou, wu}. } 
Table~\ref{table-cue-mismatch} displays the amount of cue-match and cue-mismatch sentence pairs. There are 17.4\% sentence pairs with cue-mismatch,\footnote{Note that this is only a simple approximation. We aim to demonstrate the sizeable mismatched training data rather than the accurate distribution. We manually checked 100 randomly selected sentence pairs, of which 30\% are classified incorrectly. These errors are caused by ignoring English words with negative prefixes/suffixes or viewing any Chinese words with negative characters as negative words, such as \textit{unknown} in English and \textit{nan fei (South Africa)} in Chinese.} predominantly in ZH$\rightarrow$EN, which agrees with the high amount of rephrasing we observed in our manual evaluation (Table~\ref{table-eval-manual}). 
Such cue-mismatch sentence pairs, along with cue-match pairs, can make the learning harder and cause under-translation errors when there is no paraphrase to compensate for the dropped negation cue. Thus, one possible solution is to distill or filter training data to remove cue-mismatch sentence pairs to make the learning easier. 

\subsection{Intrinsic Investigation}

We are also interested in exploring whether NMT models can distinguish negation and non-negation tokens, and therefore conduct an intrinsic investigation on hidden states -- by computing the cosine similarity between tokens with different negation tags. 
Since NMT models can translate most negation instances correctly, we hypothesize that the hidden states are capable of distinguishing negation from non-negation tokens. We investigate hidden states from both encoders and decoders. As the hidden state in the last decoder layer is used for predicting the translation, we only explore the decoder hidden states at the $6$th layer. 
We use $Sim_{ce}$ to represent the cosine similarity between negation cues and negation events, $Sim_{cs}$ to represent the cosine similarity between negation cues and tokens belonging to negation scope, and $Sim_{co}$ to represent the cosine similarity between negation cues and non-negation tokens. 
We simply use the mean representation for tokens that are segmented into subwords.

Figure~\ref{fig:similarity} shows the cosine similarity between negation cues and events, scope, and non-negation tokens, using hidden states from encoders and decoders. $Sim_{ce}$ is substantially higher than $Sim_{cs}$, and $Sim_{cs}$ is higher than $Sim_{co}$. This result reveals that negation events are closer to negation cues compared to tokens belonging to the negation scope. We can also infer that NMT models can tell negation and non-negation tokens apart as $Sim_{co}$ is distinctly lower than $Sim_{ce}$ and $Sim_{cs}$. However, even the highest $Sim_{ce}$ is only around 0.5, which means that the representations of negation components are quite different.

\begin{table*}[htbp]
\begin{center}
\scalebox{0.85}{
\begin{tabular}{ccccccccccc}
\toprule
\multicolumn{2}{c}{}& \multicolumn{3}{c}{Cues}& \multicolumn{3}{c}{Scope}& \multicolumn{3}{c}{Events}\\
\cmidrule(l){3-5} \cmidrule(l){6-8} \cmidrule(l){9-11}
Data&Model&P&R&F1&P&R&F1&P&R&F1\\
\midrule
\multirow{3}{*}{Dev} & \citet{liu2018negpar} & 0.49\phantom{0} & 0.42 & 0.45\phantom{0} & 0.64\phantom{0} & 0.44\phantom{0} & 0.50\phantom{0} & 0.40\phantom{0} & 0.27\phantom{0} & 0.32\phantom{0} \\
& \textit{ENC} & \textbf{0.915} & \textbf{0.665} & \textbf{0.770} & \textbf{0.814} & \textbf{0.530} & \textbf{0.642} & \textbf{0.598} & \textbf{0.335} & \textbf{0.429} \\
& \textit{DEC} & 0.754 & 0.488 & 0.592 & 0.738 & 0.489 & 0.588 & 0.487 & 0.272 & 0.348\\
\addlinespace
\multirow{3}{*}{Test} & \citet{liu2018negpar} & 0.478 & 0.382 & 0.425 & 0.583 & 0.312 & 0.406 & 0.338 & 0.180 & 0.235\\
& \textit{ENC} & \textbf{0.892} & \textbf{0.581} & \textbf{0.704} & \textbf{0.743} & \textbf{0.496} & \textbf{0.595} & \textbf{0.496} & \textbf{0.285} & \textbf{0.362}\\
& \textit{DEC} & 0.686 & 0.362 & 0.474 & 0.656 & 0.456 & 0.538 & 0.470 & 0.225 & 0.304\\
\bottomrule
\end{tabular}}
\caption{\label{table:res-projection} Precision (P), recall (R), and F1 scores of the negation projection tasks in EN$\rightarrow$ZH, using NMT hidden states, comparing with the word alignment based method \cite{liu2018negpar}. \textit{ENC} represents the hidden states from the 1st encoder layer in cue projection, and represents the hidden states from the 6th encoder layer in scope/event projection. \textit{DEC} denotes the hidden states from the 6th decoder layer.} 
\end{center}
\end{table*}

In the encoder, $Sim_{ce}$, $Sim_{cs}$ and $Sim_{co}$ have the same trend that the similarity is higher in upper layers. In addition, we can tell that negation cues interact with events and scope, but also non-negation tokens. 
Compared to the negation representations from encoders, the negation representations from decoders are less distinct because they are closer to each other. $Sim_{ce}$, $Sim_{cs}$ and $Sim_{co}$ are higher when using the hidden states from the $6$th decoder layer (\textit{DEC6}) than when using the $6$th encoder layer (\textit{ENC6}). 
We attribute this to the fact that hidden states in decoders are more contextualized because they consider contextual information from both the source and the target.

\subsection{Probing NMT Models on Negation}

We have shown that NMT models can distinguish negation and non-negation tokens in the previous section, but how much information about negation has been captured by NMT models is still unclear. In this section we will investigate the ability to model negation in an extrinsic way, i.e., probing hidden states on negation in a negation projection task \cite{liu2018negpar} and a negation detection task \cite{fancellu2018neural}. 
In the negation projection task, instead of projecting English negation annotations to Chinese translations using word alignment, we use probing classifiers trained on Chinese to directly generate the negation annotations.
In the negation detection task in English, we employ simple classifiers rather than specifically designed models to detect each token. 
In brief, given a hidden state, we train classifiers to predict its negation tag, \textit{cue}, \textit{event}, \textit{scope}, or \textit{others}. 

\subsubsection{Settings} 

The probing task on negation cues is a binary classification task, the output space is $\{\textit{cue}, \textit{others}\}$, while the classifiers for event and scope are tri-class classification tasks with an output space $\{\textit{cue}, \textit{event/scope}, \textit{others}\}$, because only predicting event/scope is challenging to these classifiers. 

The probing classifiers in this section are feed-forward neural networks (MLP) with only one hidden layer, using ReLU non-linear activation. The size of the hidden layer is set to 512 and we use the Adam learning algorithm. The classifiers are trained using cross-entropy loss. Each classifier is trained on the training set for 100 epochs and tuned on the development set. We select the model that performs best (F1 score) on the development set and apply it to the test set. In addition, we train 5 times with different seeds for each classifier and report average results. We use precision, recall, and F1 score as evaluation metrics.

\begin{figure*}[htbp]
\centering
        \includegraphics[totalheight=2.3cm]{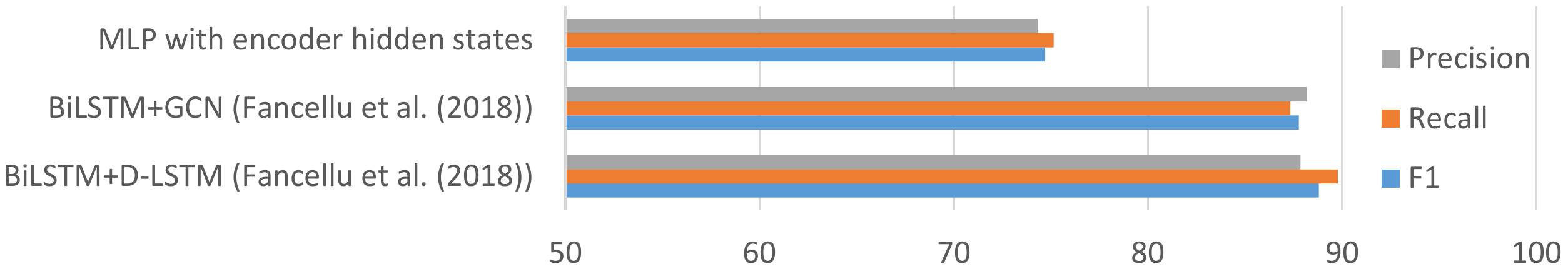}
    \caption{Results (\%) on negation scope detection in English. \textit{MLP} is the probing classifier; \textit{GCN} is graph convolutional network; \textit{D-LSTM} is bidirectional dependency LSTM. }
    \label{fig:detection}
\end{figure*}  

\begin{figure*}[htbp]
\centering
        \includegraphics[totalheight=4.2cm]{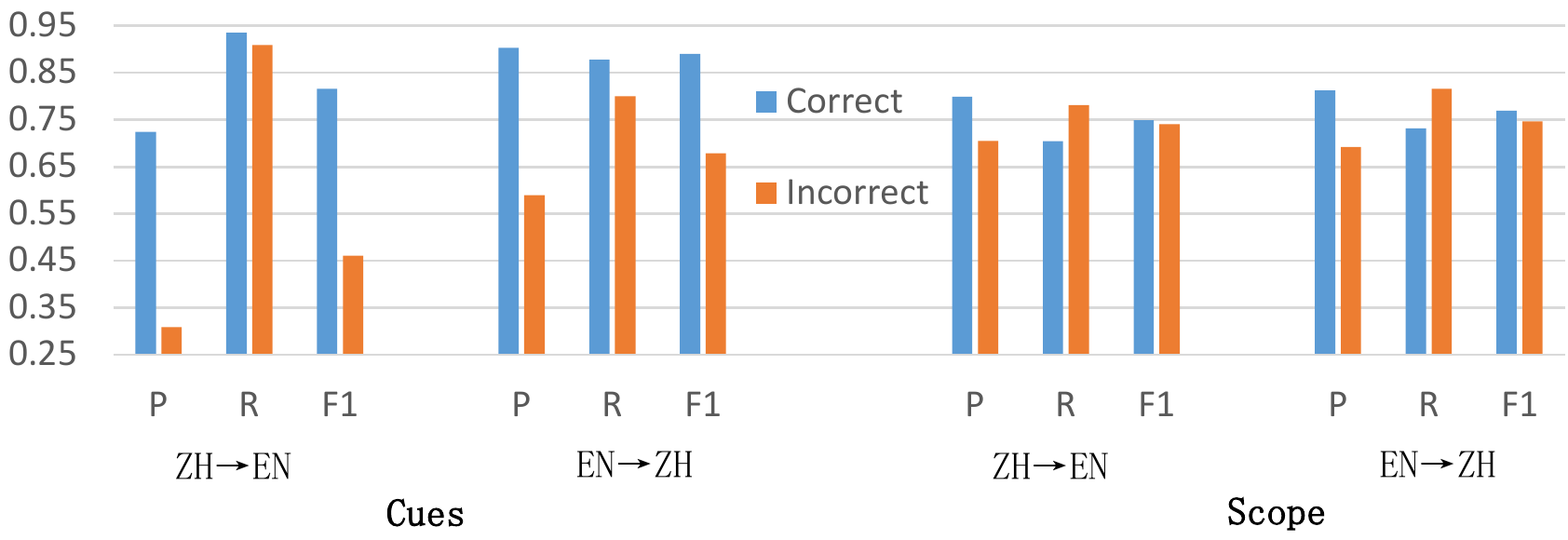}
    \caption{Results on negation cue/scope detection in ZH--EN, using encoder hidden states from sentences where the negation is correctly translated (\textit{Correct}) and incorrectly translated (\textit{Incorrect}).}
    \label{fig:right_error}
\end{figure*}  

\subsubsection{Negation Projection} 
Table~\ref{table:res-projection} shows the projection results of negation cues, scope, and events, on both development and test sets. \textit{ENC}/\textit{DEC} refers to using hidden states from encoders or decoders. \textit{ENC} achieves the best result on all the negation projection tasks and is significantly better than the word alignment based method in \citet{liu2018negpar}. \textit{ENC} also performs better than \textit{DEC}, which means that negation is better modeled in encoder hidden states than in decoder hidden states.

\begin{figure}[htbp]
\centering
        \includegraphics[totalheight=5.2cm]{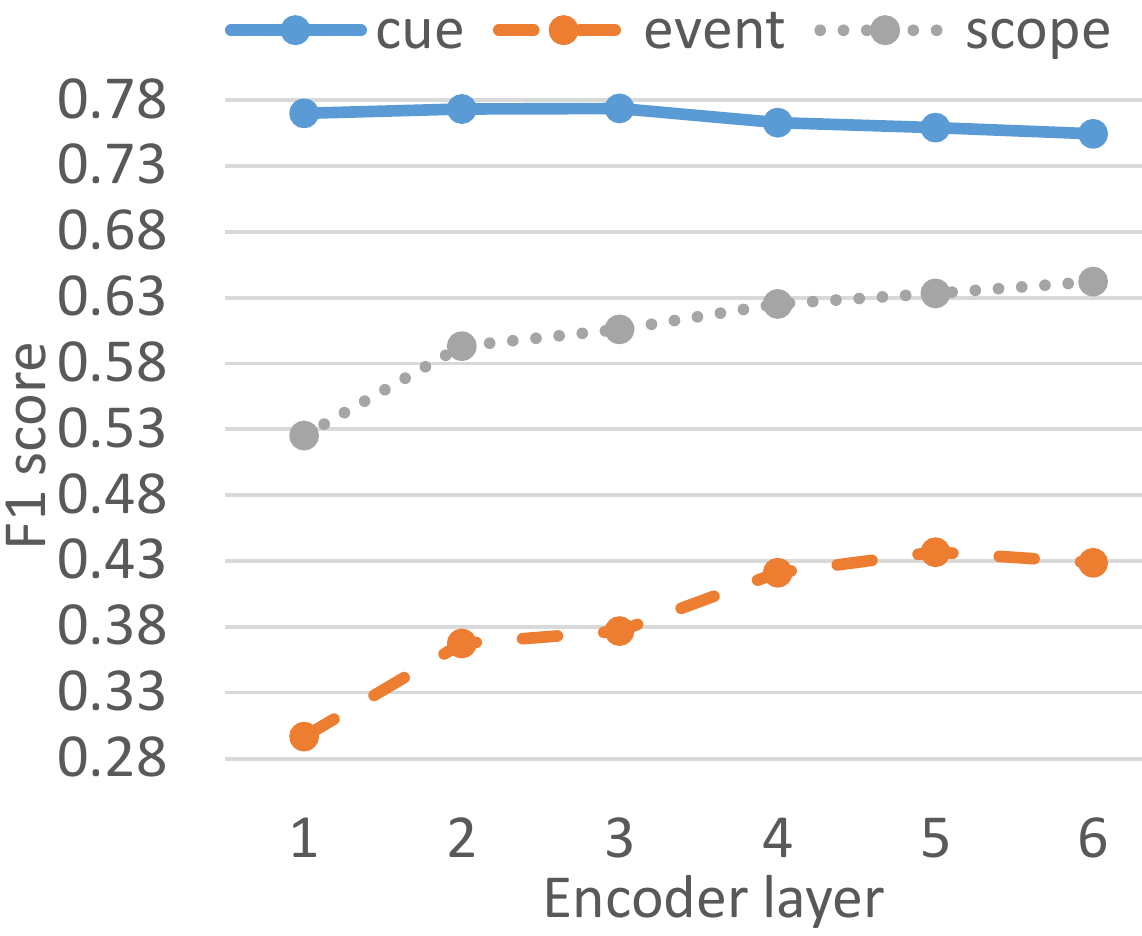}
    \caption{F1 scores of the negation projection tasks, on the development set, using hidden states from different encoder layers. }
    \label{fig:proj_layer}
\end{figure}

In addition, we investigate hidden states from different encoder layers. Figure~\ref{fig:proj_layer} shows the F1 scores on the development set, using hidden states from different encoder layers. We can see that hidden states from lower layers perform better in negation cue projection, while hidden states from upper layers are better in negation event/scope projection. One possible explanation is that negation cues in upper layers are fused with other negation information, which confuses the classifier. However, negation events/scope in upper layers interact more with negation cues and non-negation tokens, which makes them more distinctive.

\subsubsection{Negation Scope Detection}

Figure~\ref{fig:detection} shows the results of the negation scope detection task. We only report the results of using encoder hidden states that perform the best. The MLP classifier trained on encoder hidden states achieves 74.31\%, 75.14\%, and 74.72\% on precision, recall, and F1, respectively,\footnote{Here we only report the result of using hidden states from the $6$th encoder layer. We also tried hidden states from other encoder layers and decoders and got similar results as in the negation projection task.} and it is distinctly inferior to the other two models. However, methods from \citet{fancellu2018neural} are specifically designed for negation scope detection and add extra information (negation cues, POS tags) to supervise the model, while the MLP classifier is designed to jointly predict negation cues as well, only using hidden states. We can conclude that some information about negation scope is well encoded in hidden states, but there is still room for improvement.

\subsubsection{Incorrectly Translated Sentences} 

We further probe encoder hidden states from correctly and incorrectly translated sentences on negation cues and scope, to explore the quality of hidden states from incorrectly translated sentences. Note that we do not consider the under-translated cues. 
Figure~\ref{fig:right_error} exhibits the performance of negation detection on cues and scope. \textit{Correct} represents hidden states from correctly translated sentences and \textit{Incorrect} stands for hidden states from incorrectly translated sentences. \textit{Incorrect} performs worse than \textit{Correct}, especially on the negation cue detection task, which confirms the effectiveness of using probing tasks to explore the information about negation in hidden states. 

\section{Conclusion}

In this paper, we have explored the ability of NMT models to translate negation through evaluation and interpretation. 
The accuracy of manual evaluation in EN$\rightarrow$DE, DE$\rightarrow$EN, EN$\rightarrow$ZH, and ZH$\rightarrow$EN is 95.7\%, 94.8\%, 93.4\%, and 91.7\%, respectively. 
The contrastive evaluation shows that deleting a negation cue from references is more confusing to NMT models than inserting a negation cue into references, which indicates that NMT models have a bias against sentences with negation. 
We show that NMT models make fewer mistakes in EN--DE than in EN--ZH. Moreover, there are more errors in DE/ZH$\rightarrow$EN than in EN$\rightarrow$DE/ZH.

We also have investigated the information flow of negation by computing the attention weights and attention flow. 
We demonstrate that the negation information has been well passed to the decoder, and that there is no correlation between the amount of negation information transferred and whether the cues are under-translated or not. 
Thus, we consider attempts to detect or even fix under-translation of cues via an analysis or manipulation of the attention flow to have little promise. 
However, our analysis of the training data shows that negation is often rephrased, leading to cue mismatches which could confuse NMT models. This suggests that distilling or filtering training data to make grammatical negation more consistent between source and target could reduce this under-translation problem.

In addition, we show that NMT models can distinguish negation and non-negation tokens very well, and NMT models can encode substantial information about negation in hidden states but nevertheless leave room for improvement. Moreover, encoder hidden states capture more information about negation than decoder hidden states; negation cues are better modeled in lower encoder layers while negation events and tokens belonging to negation scope are better modeled in higher encoder layers. 

Overall, we show that the modeling of negation in NMT has improved with the evolution of NMT -- with deeper and more advanced networks; the performance on translating negation varies between language pairs and directions. 
We also find that the main error types on negation have shifted from SMT to NMT -- under-translation is the most frequent error type in NMT while other error types such as reordering were equally or more prominent in SMT.

We only conduct evaluation in EN--DE and EN--ZH, and German/Chinese and English are very similar in expressing negation. It will be interesting to explore languages have different characteristics on negation in the future, such as Italian, Spanish, and Portuguese, where double negation is very common.

\section*{Acknowledgments}
We thank all reviewers, action editors (Mauro Cettolo and Chris Quirk) for their valuable and insightful comments. We also thank Qianchu Liu for providing the \textit{NegPar} data set. 
We acknowledge the computational resources provided by CSC in Helsinki and Sigma2 in Oslo through NeIC-NLPL (www.nlpl.eu). 
GT was mainly funded by the Chinese Scholarship Council (NO. 201607110016).

\bibliography{negation}
\bibliographystyle{acl_natbib}

\end{document}